\documentclass[10pt,twocolumn,letterpaper]{article}

\usepackage[pagenumbers]{cvpr} % To force page numbers, e.g. for an arXiv version
\definecolor{cvprblue}{rgb}{0.21,0.49,0.74}
\usepackage[colorlinks,linkcolor=black,citecolor=blue,urlcolor=blue]{hyperref}

\def\confName{CVPR}
\def\confYear{2026}

\title{Exploring the Role of Synthetic Data Augmentation in Controllable Human-Centric Video Generation}

\author{
Yuanchen Fei$^{*}$\\
Hunan University\\
Westlake University
\and
Yude Zou$^{*}$\\
Shanghai Jiaotong University
\and
Zejian Kang\\
Westlake University\\
Zhejiang University
\and
Ming Li\\
Zhejiang University\\
Westlake University\\
Shanghai Innovation Institute
\and
Jiaying Zhou\\
Sun Yat-Sen University
\and
Xiangru Huang\\
Westlake University
}

\begin{document}
\maketitle
\begin{abstract}
Controllable human video generation aims to produce realistic videos of humans with explicitly guided motions and appearances, serving as a foundation for digital humans, animation, and embodied AI. However, the scarcity of large-scale, diverse, and privacy-safe human video datasets poses a major bottleneck, especially for rare identities and complex actions. Synthetic data provides a scalable and controllable alternative, yet its actual contribution to generative modeling remains underexplored due to the persistent Sim2Real gap. In this work, we systematically investigate the impact of synthetic data on controllable human video generation. We propose a diffusion-based framework that enables fine-grained control over appearance and motion while providing a unified testbed to analyze how synthetic data interacts with real-world data during training. Through extensive experiments, we reveal the complementary roles of synthetic and real data and demonstrate how properly selected synthetic samples can enhance motion realism, temporal consistency, and identity preservation. Our study offers the first comprehensive exploration of synthetic data’s role in human-centric video synthesis and provides practical insights for building data-efficient and generalizable generative models.
\end{abstract}    
\section{Introduction}
\label{sec:introduction}

%Please follow the steps outlined below when submitting your manuscript to the IEEE Computer Society Press.
%This style guide now has several important modifications (for example, you are no longer warned against the use of sticky tape to attach your artwork to the paper), so all authors should read this new version.

%-------------------------------------------------------------------------
Controllable human video generation has become an essential research direction in computer vision and generative modeling, enabling precise manipulation of motion, appearance, and temporal dynamics in synthesized videos~\cite{zhou2024realisdance,zhang2024mimicmotion,men2025mimo,zhu2025generative}. Such capability is crucial for a wide range of applications, including digital human animation~\cite{jang2024faces,gao2023high}, film production, virtual reality, and human–computer interaction. Controllable video generation aims to synthesize videos that not only appear realistic and coherent, but also strictly follow given control signals such as motion, pose, or identity~\cite{hu2025animate,hu2024animate,xu2024magicanimate}. Achieving such controllability enables fine-grained human animation, personalized video synthesis, and the ability to simulate rare or unseen actions, all of which are highly valuable in both research and industry.

However, training high-quality controllable video generation models remains challenging. Human video datasets are difficult and costly to collect, as they require diverse subjects, accurate motion capture, and strict privacy protection. As a result, existing models often struggle to generalize to rare motions, unseen identities, or fine-grained appearance variations, limiting their scalability and performance. Synthetic (simulated) data, which can be programmatically generated using computer graphics or motion capture systems, offers a promising solution to this data scarcity problem~\cite{wang2025exploring,sariyildiz2023fake,zhang2025accvideo}. Synthetic videos can be produced at scale with precise annotations and controllable variations in motion, lighting, and viewpoint—features that are difficult to achieve in real-world data collection.

Nevertheless, a major obstacle persists: the simulation-to-reality gap (Sim2Real gap). Synthetic and real data differ significantly in visual distribution, texture fidelity, and physical realism. These discrepancies often lead to degraded performance when models trained on synthetic data are applied to real-world scenarios. Although synthetic data has been successfully employed in tasks such as object detection, robotic control, and reinforcement learning~\cite{hofer2021sim2real,richter2016playing,tremblay2018training}, its potential in human-centric video generation remains largely unexplored. Bridging this gap is therefore crucial to fully exploit synthetic data for controllable video synthesis.

\begin{figure*}[t]
    \centering
    \includegraphics[width=0.8\textwidth]{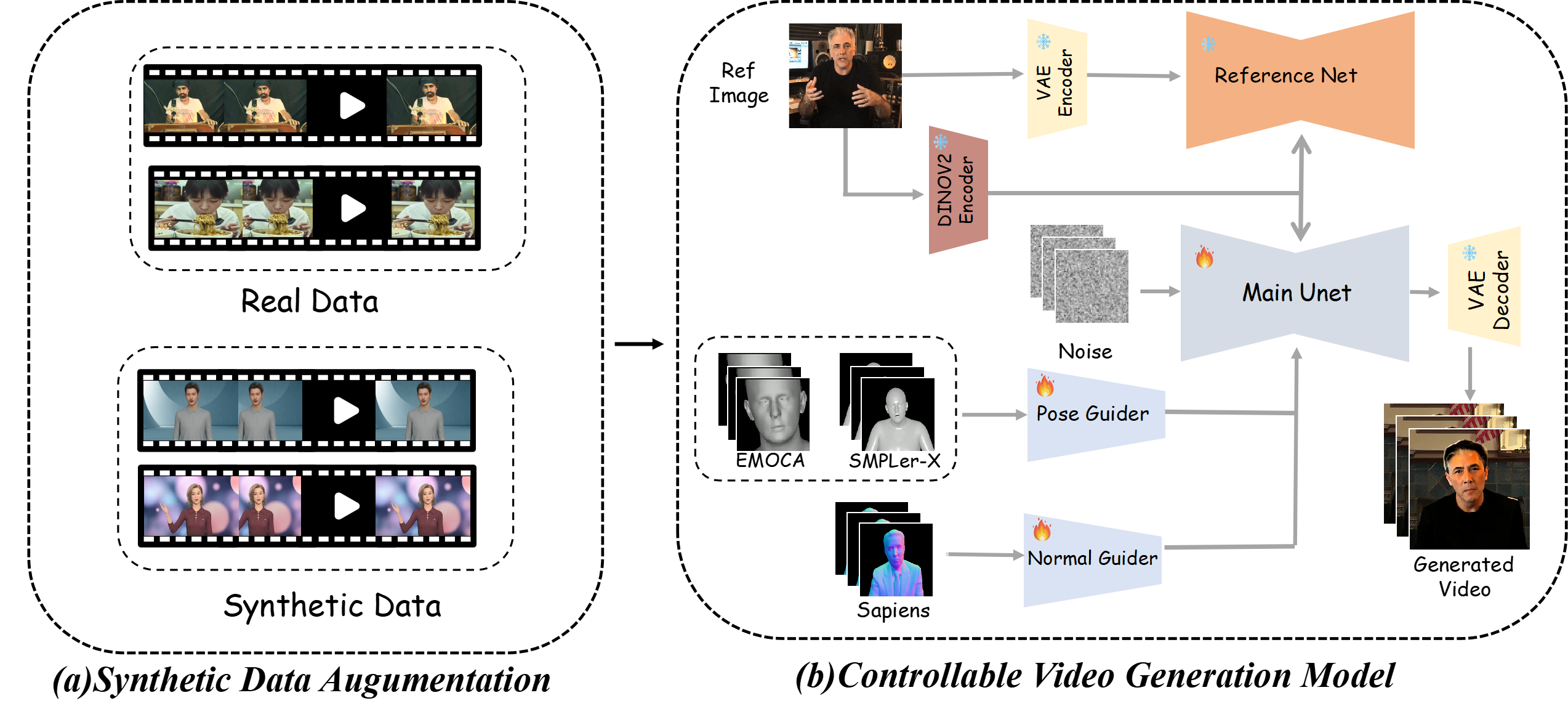}
    \caption{The overview of our work. We present (a) a comprehensive exploration of synthetic data augmentation on (b) our controllable human-centric video generation model.}
    \label{fig1}
\end{figure*}

To address this challenge, we present the first systematic exploration of synthetic video data in controllable video generation. Our framework effectively bridges the Sim2Real gap, enabling synthetic videos to positively contribute to the training of real-world video generation models. By integrating simulated motion data with real video distributions, our approach allows the model to learn more robust and generalizable motion representations while preserving identity consistency. The overview of our work is shown in Fig~\ref{fig1}. Extensive experiments demonstrate that our method achieves significant improvements in multiple metrics, validating both the feasibility and effectiveness of incorporating synthetic data into video generation pipelines.

In summary, our main contributions are as follows:

(1) We propose a new framework capable of synthesizing realistic human videos driven by explicit and interpretable control signals. Our approach effectively integrates both appearance and motion control, enabling fine-grained manipulation of human attributes in a coherent and visually consistent manner.

(2) We are the first, to the best of our knowledge, to systematically explore the role of synthetic data in training video generation models. By incorporating diverse types of simulated data during training, we investigate the synergistic relationship between synthetic and real data, providing valuable insights into how simulation data can enhance model generalization and controllability.

(3) We further study how domain-specific synthetic data can improve generation performance in corresponding tasks. Based on our analysis, we propose possible strategies for selecting and utilizing synthetic datasets that most effectively complement real data, offering a fresh perspective on mitigating Sim2Real gap and leveraging simulation data to boost controllable video generation.

\section{Related Work}
\label{sec:related_work}

\paragraph\noindent\textbf{Diffusion Model for Video Generation.} Diffusion-based generative models have become a dominant approach for high-fidelity image synthesis and have been extended rapidly to video generation. The foundational theory and training recipe for this family of models were established in the context of image generation by denoising diffusion probabilistic models (DDPM)\cite{ho2020denoising} and score-based SDEs\cite{song2020score}, which show how to parameterize and learn reverse-noising processes to produce high-quality samples. Unlike image generation, video generation requires the diffusion model to have consistency and smoothness in the temporal dimension. Currently, the mainstream method for video generation models is to add a temporal layer to the image generation diffusion model to maintain temporal consistency\cite{ho2022video}. The latent diffusion model proposes a denoising technique in the latent space\cite{rombach2022high}, which maintains the fidelity of the model while improving the computational efficiency.

Later, in the research of controllable video generation applications based on the diffusion model, controlnet\cite{zhang2023adding} innovatively proposed a control method of injecting unet containing identity information into the main diffusion unet, which promoted a major breakthrough in video generation towards greater controllability and accuracy. The method based on adding control conditions to the diffusion model has also been widely adopted in subsequent controllable video generation work.

\paragraph\noindent\textbf{Controllable Human-Centric Video Generation.} The controllable generation of human videos aims to synthesize temporally coherent and identity-preserving video sequences from one or more static reference images under motion or text conditioning. Early studies relied primarily on generative adversarial networks (GANs)\cite{goodfellow2020generative,wang2018video} for pose transfer and image animation. With the emergence of diffusion models (DMs), human video generation has undergone a significant paradigm shift. Diffusion-based methods leverage the strong denoising and representation capabilities of pretrained text-to-image models while extending them to the spatiotemporal domain.

Recent advances have explored diverse conditioning strategies and network architectures to improve motion fidelity and appearance alignment. DisCo\cite{wang2024disco} introduces dual ControlNets to separately handle pose and background for fine-grained disentanglement. Animate Anyone\cite{hu2024animate} extends AnimateDiff\cite{guo2023animatediff} with a ReferenceNet for appearance extraction and a lightweight pose guider for structural control. MagicAnimate\cite{xu2024magicanimate} employs DensePose\cite{neverova2018dense}-conditioned ControlNet to enhance temporal coherence and body realism. Building upon these, RealisDance\cite{zhou2024realisdance} leverages a 1D temporal U-Net and a reference encoder for identity-preserving dance video synthesis. MimicMotion\cite{zhang2024mimicmotion} decouples global appearance and local motion through dual-branch feature decomposition, improving controllability. MIMO\cite{men2025mimo} models body parts independently in latent space for spatially decomposed character synthesis. StableAnimator\cite{tu2025stableanimator,tu2025stableanimator++} augments Stable Diffusion with temporal attention and motion adapters, achieving flexible motion control while preserving identity fidelity.

However, progress in controllable human video generation remains constrained by the difficulty of acquiring large-scale, diverse, and privacy-compliant human motion datasets, which limits model generalization across poses, clothing, and environments. Moreover, maintaining long-term identity consistency under complex motion and temporal variation continues to be a major challenge.

\paragraph\noindent\textbf{Synthetic Data for Model Training.} Synthetic data has been widely applied in fields such as computer vision, robotics, and reinforcement learning.\cite{hofer2021sim2real,richter2016playing,tremblay2018training} It is generated through computer graphics and can automatically provide precise annotations, thereby alleviating the problem of obtaining real data and providing low-cost, large-scale, and diverse samples for model training. For example, Virtual KITTI\cite{cabon2020virtual} generates pixel-level labels by cloning real urban driving scenes, demonstrating the good transferability of synthetic data in tasks such as detection, segmentation, tracking, and optical flow. \cite{sadeghi2016cad2rl} trained an unmanned aerial vehicle obstacle avoidance strategy in a fully synthetic three-dimensional environment, achieving real flight without real images.

However, there are visual distribution and physical modeling differences between synthetic and real data, resulting in a decline in transfer performance from simulation to reality (Sim2Real)\cite{chigot2025style}. Visually, synthetic images do not match real scenes in terms of lighting, materials, and texture details; physically, \cite{tobin2017domain} pointed out that the simulation environment is difficult to precisely reproduce details such as non-rigid deformation, friction wear, fluid dynamics, and sensor noise. These unmodeled differences constitute a "reality gap" (reality gap), which becomes the main bottleneck for Sim2Real transfer. Therefore, how to narrow the domain differences between simulation and reality remains a key research direction for improving model generalization ability. Some fields have already conducted research on using synthetic data for model training, such as in the image classification task, where \cite{wang2025exploring} quantitatively studied the equivalence relationship between open-set and closed-set generated image datasets and real datasets on the performance of classification models. However, these studies on synthetic data have not yet extended to the video generation domain. Our work aims to study the impact and role of synthetic data in the training of controllable video generation models and explore whether the sim2real gap between synthetic data and real data in videos can be bridged.

\section{Preliminary}

\subsection{Problem Setup}

We focus on controllable human-centric video generation, a task that aims to synthesize temporally coherent human videos conditioned on explicit control signals. Formally, given a reference image the target identity, and a sequence of control inputs representing body pose, facial expression, or other motion cues, the model generates a video that preserves the reference appearance while following the prescribed motion. In our work, This controllable setup provides a unified platform for studying synthetic data augmentation, enables controlled experiments on how synthetic data of different composition, realism, or semantic similarity influence model generalization—particularly in bridging the Sim2Real gap.

\subsection{Model Architecture}

Our proposed model is a controllable video generation framework that leverages multiple control signals to generate high-quality human dance videos. The overall architecture can be generalized in Figure~\ref{fig1} (b). It consists of four main components: a reference encoder, a main denoising network, pose guidance modules, and a CLIP-based appearance encoder.

The model takes as input a reference image $\mathbf{I}_{\mathrm{ref}}\in\mathbb{R}^{3\times H\times W}$ and a sequence of control signals, including SMPLer-X~\cite{loper2023smpl,cai2023smpler} body meshes $\mathbf{S}\in\mathbb{R}^{3\times F\times H\times W}$ and facial expression maps $\mathbf{H}\in\mathbb{R}^{3\times F\times H\times W}$ from EMOCA~\cite{danvevcek2022emoca}
. Since the motion representations from SMPLer-X provide an accurate modeling of body dynamics but lack fine-grained hand articulation, we additionally incorporate surface normal maps $\mathbf{N}\in\mathbb{R}^{3\times F\times H\times W}$ from Sapiens~\cite{khirodkar2024sapiens} to supplement detailed geometric cues, especially for hand and local surface movements, where $\mathbf{F}$ denotes the number of frames. The model generates a video sequence $\mathbf{V}\in\mathbb{R}^{3\times F\times H\times W}$ that matches the given control signals while preserving the appearance of the reference image.

The generation process can be formulated as:
\begin{equation}
\mathbf{V}=\mathcal{G}(\mathbf{I}_{\mathrm{ref}},\mathbf{S},\mathbf{H},\mathbf{N};\theta)
\end{equation}
where $\mathcal{G}$ is the generator parameterized by $\theta$.

\paragraph\noindent\textbf{Reference and Appearance Encoding.} To preserve the appearance information from the reference image, we employ a ReferenceNet attention mechanism~\cite{xu2024magicanimate}. The reference encoder consists of a frozen UNet2D model that processes the reference image~\cite{kingma2013auto}:
\begin{equation}
\mathbf{z_{ref}}=\mathrm{VAE_{enc}}(\mathbf{I_{ref}})
\end{equation}
where $\mathbf{z}_{\mathrm{ref}}\in\mathbb{R}^{4\times h\times w}$ is the latent representation with h=H/8,w=W/8 for SD 1.5 architecture.

The ReferenceNet attention operates in a write-read paradigm. During the forward pass, the reference encoder writes spatial features at each transformer block into a memory bank. The main denoising network then reads from this bank through cross-attention mechanisms:
\begin{equation}
\mathrm{Attn}(\mathbf{Q},\mathbf{K}_\mathrm{ref},\mathbf{V}_\mathrm{ref})=\mathrm{softmax}\left(\frac{\mathbf{Q}\mathbf{K}_\mathrm{ref}^T}{\sqrt{d}}\right)\mathbf{V}_\mathrm{ref}
\end{equation}
where $\mathbf{Q}$ are queries from the main network, and $\mathbf{K}_{\text{ref}}$, $\mathbf{V}_{\text{ref}}$ are keys and values from the reference encoder. To maintain temporal consistency, we set the timestep for the reference encoder to zero, ensuring stable feature extraction across all generation steps.
For high-level appearance semantics, a frozen DINOv2 encoder~\cite{oquab2023dinov2} extracts semantic embeddings:
\begin{equation}
\mathbf{c}_{\mathrm{clip}}=\mathrm{DINOv}2(\mathbf{I}_{\mathrm{ref}}),
\end{equation}
which are projected via a residual feed-forward network with GEGLU~\cite{shazeer2020glu} activation:
\begin{equation}
\mathbf{c}_{\mathrm{proj}}=\mathrm{FFN}(\mathbf{c}_{\mathrm{clip}})+\mathbf{W}\mathbf{c}_{\mathrm{clip}}.
\end{equation}
These embeddings provide identity-consistent conditioning to all UNet cross-attention layers.

\paragraph\noindent\textbf{Dual Pose Guidance Modules.} We design two separate pose guidance modules to handle different types of control signals:

(a) Pose Guider. This module fuses SMPL body mesh and facial expression maps:
\begin{equation}
\mathbf{p}_{\mathrm{body}}=\text{PoseGuider}(\mathbf{S},\mathbf{H}),
\end{equation}
where an adaptive gating mechanism controls the contribution of each input:
\begin{equation}
\begin{split}
\mathbf{g} &= \sigma(\mathrm{Conv}([\mathbf{S}, \mathbf{H}])) , \\
\mathbf{p}_{\mathrm{body}} &= \mathbf{g} \odot \mathrm{Backbone}([\mathbf{S}, \mathbf{H}]).
\end{split}
\end{equation}

(b) Normal Guider. To inject fine-grained geometry, the normal guider produces compact 1D descriptors:
\begin{equation}
\mathbf{p}_{\mathrm{normal}}=\mathrm{Linear}(\mathrm{Flatten}(\mathrm{Backbone}(\mathbf{N})))\in\mathbb{R}^{1\times512}.
\end{equation}
These features modulate the video UNet via temporal cross-attention:
\begin{equation}
\mathbf{h}^{\prime}=\mathrm{Attn}(\mathbf{h},\mathbf{p}_{\mathrm{normal}},\mathbf{p}_{\mathrm{normal}})+\mathbf{h}.
\end{equation}

\paragraph\noindent\textbf{Main Denoising Network.} The main denoising network adopts a UNet3D backbone for video generation (stage 2) and a UNet2D for image generation (stage 1). For the video generation stage, we follow the temporal module design from AnimateDiff~\cite{guo2023animatediff}:
\begin{equation}
\mathbf{h}_t = \text{SpatialAttn}(\mathbf{h}_{t-1}) + \text{TemporalAttn}(\mathbf{h}_{t-1}),
\end{equation}
where spatial and temporal attention are applied sequentially. Temporal attention operates along the frame dimension. The network input consists of 12 channels: 4 for the noisy latent \(\mathbf{z}_t\), 4 for the background latent \(\mathbf{z}_{\text{bg}}\) derived from segmentation masks provided by Grounded-SAM2~\cite{ren2024grounded}, and 4 for the foreground mask latent \(\mathbf{z}_{\text{fg}}\). This design enables compositional generation by disentangling human motion from static background content.

The denoising objective follows the v-prediction formulation~\cite{salimans2022progressive} with zero-SNR noise scheduling~\cite{lin2024common}:
\begin{equation}
\mathcal{L} = \mathbb{E}_{t,\mathbf{z}_0,\epsilon}\left[w(t)\,\|\mathbf{v}_\theta(\mathbf{z}_t, t, \mathbf{c}) - \mathbf{v}_t\|_2^2\right],
\end{equation}
where \(w(t)=\dfrac{\min(\text{SNR}(t),\gamma)}{\text{SNR}(t)+1}\) and \(\gamma=5.0\).

\subsection{Two-Stage Training and Inference.}
\subparagraph{Stage 1: Image-Level Pretraining.} We train a 2D UNet to align control signals with appearance under a fixed reference timestep. This stage establishes spatial understanding and identity preservation.
\subparagraph{Stage 2: Video-Level Finetuning.} We extend the model to 3D UNet with temporal attention layers initialized from AnimateDiff.
During training, we freeze spatial layers from Stage 1 and update only temporal motion modules.
Pose shuffling and random control-signal dropout improve temporal robustness.
\subparagraph{Inference.} We apply DDIM sampling\cite{song2020denoising} for 50 steps:
\begin{multline}
\mathbf{z}_{t-1}=\sqrt{\alpha_{t-1}}
\left(\frac{\mathbf{z}_t-\sqrt{1-\alpha_t}\epsilon_\theta(\mathbf{z}_t,t,\mathbf{c})}{\sqrt{\alpha_t}}\right) \\
+\sqrt{1-\alpha_{t-1}}\epsilon_\theta(\mathbf{z}_t,t,\mathbf{c}).
\end{multline}

The decoded latent yields the final video:
\begin{equation}
\mathbf{V}=\mathrm{VAE}_{\mathrm{dec}}(\mathbf{z}_0).
\end{equation}

\begin{figure}[t]
    \centering
    \includegraphics[width=0.55\textwidth]{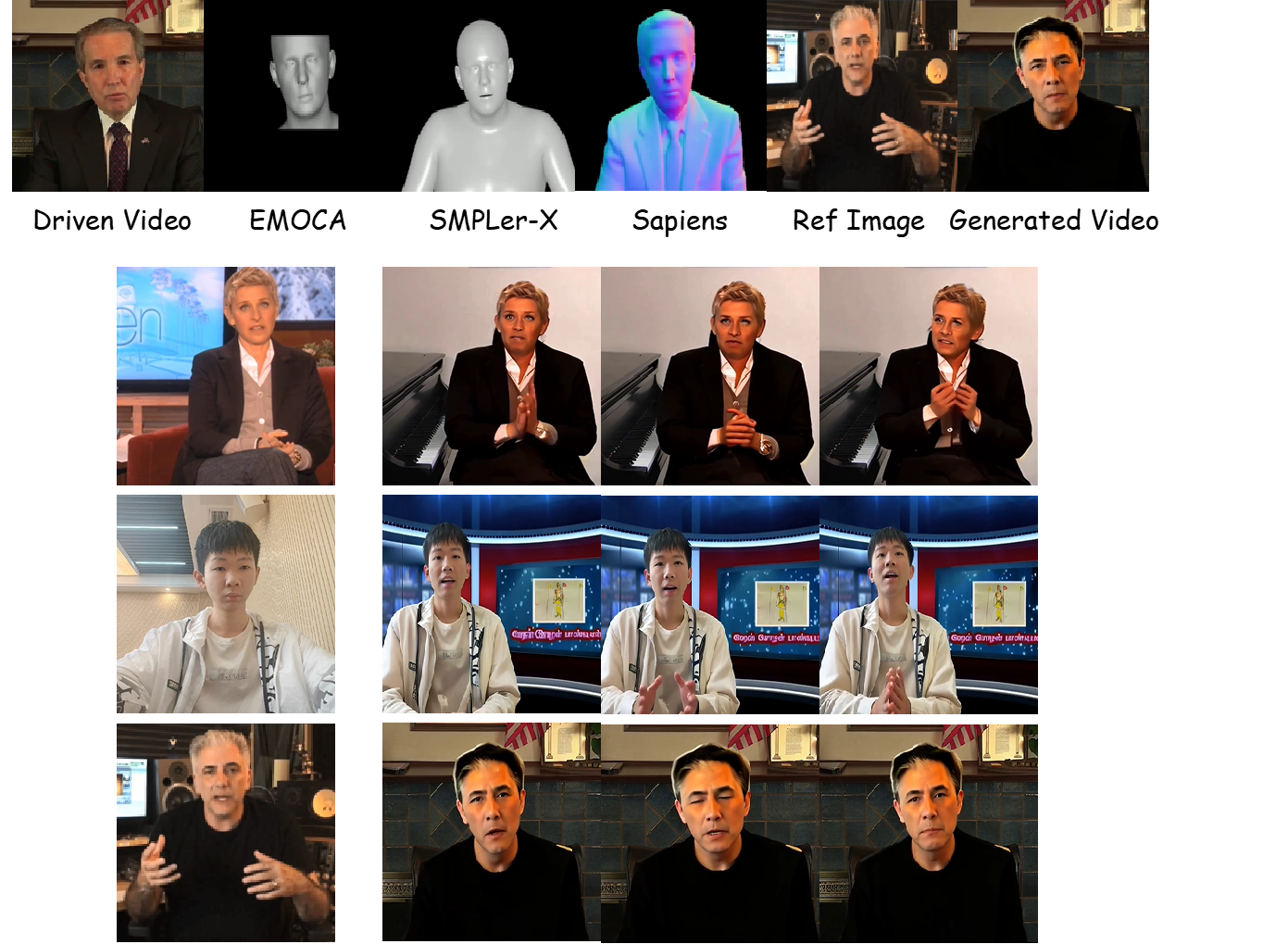}
    \caption{Inference results and control signals. Our model achieves consistent character animation through various control signals while enabling the integration and interaction between characters and their environments.}
    \label{fig2}
\end{figure}

By jointly integrating multi-source control signals, semantic embeddings, and geometric priors, our framework achieves precise control over both appearance and motion. The inference performance of our model is shown in Figure~\ref{fig2}. Unlike prior video diffusion models that rely solely on pose or optical flow, our dual-guided design enables better disentanglement between spatial style and temporal dynamics, providing a more generalizable foundation for controllable human video synthesis.

\section{Method}

Obtaining large-scale, high-quality human video data remains a significant challenge, particularly for rare actions or identities. This scarcity limits the performance of controllable video generation models, as they struggle to generalize to underrepresented scenarios. Synthetic data, generated in a controllable manner, provides a promising solution to this problem: it can be produced in large quantities, covers diverse conditions, and allows precise control over actions and appearance. In this work, we systematically investigate whether and how synthetic data can improve controllable video generation performance.

In this work we systematically study whether, and under which conditions, synthetic data can improve controllable video generation. Concretely, we perform three sets of experiments to (i) test fine-tuning a mature real-data model on synthetic data, (ii) examine progressive synthetic-data augmentation from scratch in a narrow domain, and (iii) evaluate targeted selection of synthetic samples based on semantic similarity to specific test videos.

\begin{figure}[t]
    \centering
    \includegraphics[width=0.45\textwidth]{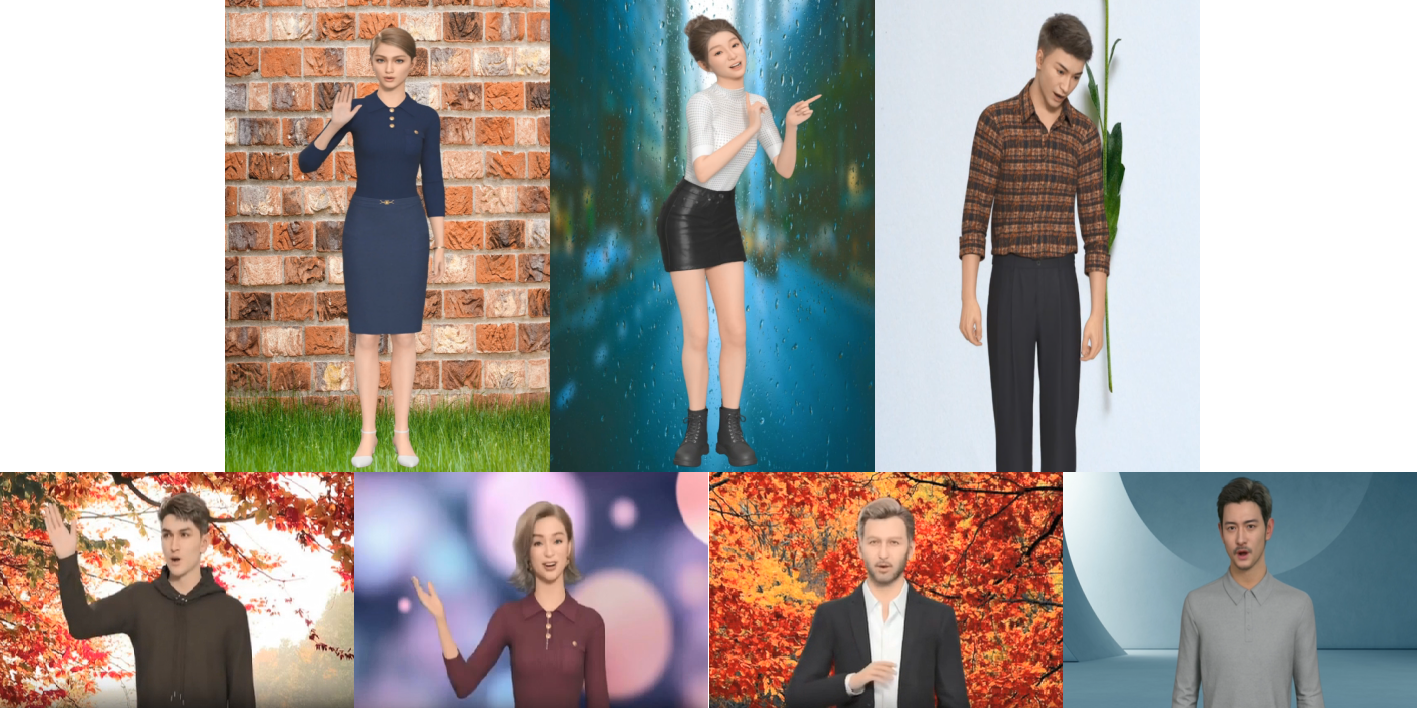}
    \caption{Visualization of the synthetic human video data used in our work. Our synthetic human videos are visually faithful, motion-diverse, and temporally coherent, providing effective data augmentation that boosts controllable video generation performance.
}
    \label{fig5}
\end{figure}

\paragraph\noindent\textbf{Fine-Tuning a model on Synthetic Data.} 
To systematically analyze the influence of synthetic fine-tuning, we construct a large-scale real video dataset collected from Bilibili, covering a wide range of speakers, identities, and motion patterns. Using this dataset, we train a baseline model that achieves strong performance. In addition, we generate a complementary synthetic dataset (Figure~\ref{fig5}), containing paired motion–appearance sequences with consistent temporal structure.

We then explore how synthetic data can be incorporated into the training pipeline under two different configurations: (A) Training is conducted purely on real data, (B) Training includes fine-tuning on synthetic data.
Both variants are evaluated on the SpeakerVid-5M~\cite{zhang2025speakervid} dataset which contains a number of individual half-body speech videos, ensuring a consistent real-world testing environment. This design enables us to isolate the influence of synthetic fine-tuning across different training phases.

The results show a consistent improvement in visual fidelity and temporal coherence after synthetic fine-tuning, confirms that synthetic data can serve as a meaningful complement to real data and motivates a deeper exploration of its potential benefits.

\paragraph\noindent\textbf{Scaling Synthetic Data for Balanced Learning.} Having established that synthetic data can enhance pretrained models, we next examine whether “more synthetic data” necessarily leads to “better performance”. We conduct controlled experiments within a constrained domain of single-person upper-body speech videos. Starting from a small subset of real samples from AVSpeech~\cite{ephrat2018looking} dataset which perfectly fits our domain, we progressively augment it with synthetic data at different ratios: 0:1, 1:1, 2:1, 4:1, and 8:1 (synthetic:real), training all models from scratch under identical configurations.

The results reveal two important insights. First, model performance generally improves as synthetic data increases, confirming that synthetic augmentation can effectively enhance training efficiency and generalization. Second, this trend is not strictly monotonic. Excessive or low-quality synthetic data introduces noise and degrades specific metrics. These observations suggest that not all synthetic samples are equally beneficial, and highlight the importance of selecting or generating synthetic data that aligns closely with the target distribution to bridge the Sim2Real gap. This insight directly motivates our third experiment.

\paragraph\noindent\textbf{Targeted Synthetic Data Selection.} 
Building on the findings above, we design the third experiment to explore how targeted selection of synthetic data can further improve performance and mitigate the Sim2Real gap. 

Specifically, we focus on a set of unseen real-world test videos and aim to identify the most relevant synthetic samples for augmentation. Using CLIP~\cite{radford2021learning} feature embeddings, we compute similarity between candidate synthetic videos and the target test videos, and select the top-n samples for augmentation.
Let $\mathbf{v}_{t}$ denote the CLIP embedding of a target test video, and let $\mathbf{v}_{i}$ denote embeddings of synthetic samples. The similarity score is computed as the cosine similarity:
\begin{equation}
S_i=\frac{\mathbf{v}_t^\top\mathbf{v}_i}{\|\mathbf{v}_t\|\|\mathbf{v}_i\|}
\end{equation}
We rank all synthetic videos by $S_{i}$ and select the top-n samples with highest similarity as augmentation candidates. This ensures that the selected synthetic data shares strong semantic and visual correspondence with the target domain.

We compare three settings: (A) random selection, (B) manual selection, and (C) CLIP-based semantic selection. All groups share the same amount of selected data. Quantitative results show that CLIP-based selection yields smoother motion, better identity consistency, and higher overall realism than both random and manual selection. This demonstrates that semantically aligned synthetic data exhibits improved performance on designated test set performance. Moreover, it further reveals that when facing challenges such as rare identities or unseen motion patterns, generating semantically similar synthetic data proves to be a practical and efficient strategy for targeted performance improvement and mitigating the Sim2Real gap.

\section{Experiments}
\label{sec:experiments}

%Our experiment aims to explore to what extent synthetic data in a specific domain can enhance the performance of model training. The dataset we collect focused on the domain of single-person half-body speech presentations. The real-person dataset comes from the AVSpeech dataset(uniformly rescaled to a resolution of 960x540), while the synthetic dataset is from Mofayouyan.

\subsection{Training Settings}

We initialize our Denoising UNet from the realistic-vision-v51 checkpoint and employ DINOv2  as our image encoder for reference image feature extraction. Our training pipeline consists of two distinct stages with different optimization objectives and trainable parameters.

Training Data and Resolution. Each video frame is accompanied by multi-modal pose representations including SMPL body meshes, EMOCA emotional representations, and surface normal maps, providing comprehensive human body guidance. During training, reference images are extracted using random frame selection, to ensure robust identity preservation across different scenarios.

In the first stage, we focus on establishing strong spatial pose-to-image mapping. All network parameters are trainable, including the UNet, Pose Guider, and CLIP projector. We use a batch size of 12 and %train for 30,000 iterations 
with a learning rate of 1e-4 using the AdamW~\cite{kingma2014adam} optimizer. This stage operates on individual frames (image\_finetune=True), enabling the model to learn precise pose control without the complexity of temporal modeling. %The learning rate follows a constant schedule with 5,00 warmup steps. 
To enhance training robustness, we randomly drop individual pose modalities (emoca, smpl, normal) with a $1\%$ probability each, encouraging the model to learn complementary information from multiple pose representations.

In the second stage, we freeze the spatial components (Denoising UNet, Pose Guider, and CLIP projector) and exclusively train the motion modules to learn temporal dynamics. The motion modules are initialized with pre-trained motion modeling weights (mm\_v2) and inserted at resolutions [1, 2, 4, 8] within the UNet architecture. Each motion module consists of temporal self-attention layers with 8 attention heads and temporal position encoding (maximum length=32). %We employ zero initialization for the output projection layers to ensure stable training initialization.z这个点好像method介绍模型架构时说过了
Training uses 16-frame video clips sampled at 8 fps with a batch size of 1, %running for 15,000 iterations 
with a reduced learning rate of 5e-5. To improve temporal robustness, we apply pose shuffle augmentation with a $5\%$ probability, randomly permuting frames within the pose sequence to prevent overfitting to specific motion patterns.

Training Infrastructure. %All training is conducted on 1 NVIDIA A100 GPU. 
We employ mixed-precision training (FP16) to reduce memory consumption and accelerate computation. The VAE encoder is frozen throughout both stages to maintain consistent latent space properties.
%We conducted five sets of experiments to verify the improvement of the generated quality of the model through the exponential addition of synthetic data to the dataset. Specifically, we trained the model on datasets with only 210 real data, 210 synthetic data + 210 synthetic data, 420 synthetic data + 210 synthetic data, 840 synthetic data + 210 synthetic data, and 1680 synthetic data + 210 synthetic data, respectively, and evaluated the generated quality using a 22-real data set. We evaluate the quality of a single frame image, with the indicators including L1 error, Structural Similarity Index (SSIM) [55], Learned Perceptual Image Patch Similarity (LPIPS), and Peak Signal-to-Noise Ratio (PSNR).
%Additionally, we conducted another set of experiments by adding specific action synthetic data to the dataset (taking hand raising as an example), using real hand-raising action videos as verification, to assess whether there is an improvement effect after adding specific action synthetic data to the dataset compared to before.
\subsection{Experiments Settings}
\paragraph\noindent\textbf{Synthetic Fine-Tuning on a Pretrained Baseline.} We first train a strong controllable human video generation baseline on the Bilibili dataset. The first stage is trained for 300k steps on four NVIDIA A100 GPUs, and the second stage is trained for 100k steps on three NVIDIA A100 GPUs.
To analyze the contribution of synthetic data at different stages, we design the following configurations: directly evaluated using the pretrained model trained purely on real videos (Baseline), and fine-tuned the baseline model on the synthetic dataset for 30k steps on a single A100 GPU (Finetuned).

Both models are evaluated on SpeakerVid-5M using only real human videos as test data to assess real-world generalization.
\vspace{-0.1in}\paragraph\noindent\textbf{Scaling Synthetic-to-Real Ratios.}
In the second set of experiments, we restrict the domain to single-person upper-body speech videos to conduct controlled analysis. The real videos are sampled from AVSpeech, while the synthetic videos are collected from our simulated dataset.
We train five models with different synthetic-to-real data ratios: 0:1, 1:1, 2:1, 4:1, and 8:1, respectively. All models are trained from scratch with identical hyperparameters — 30k steps for stage 1 and 15k steps for stage 2, each on a single NVIDIA A100 GPU. Evaluation is performed on real videos from SpeakerVid-5M.
\paragraph\noindent\textbf{Methods of Targeted Screening on Synthetic Data}
In the third experiment, we investigate targeted synthetic data selection for domain adaptation. We fix a set of wild real videos as test data and construct three training groups, each containing 210 real videos sampled from AVSpeech. We augment each group with n = 50 synthetic samples selected via different strategies: (1) Random selection, (2) Manual selection by human inspection, (3) CLIP-based semantic similarity selection.

All models are trained from scratch with identical schedules (30k steps for stage 1, 15k steps for stage 2, both on a single NVIDIA A100 GPU) and tested on the same wild real video set.

\subsection{Comparisons}

\begin{figure}[t]
    \centering
    \includegraphics[width=0.45\textwidth]{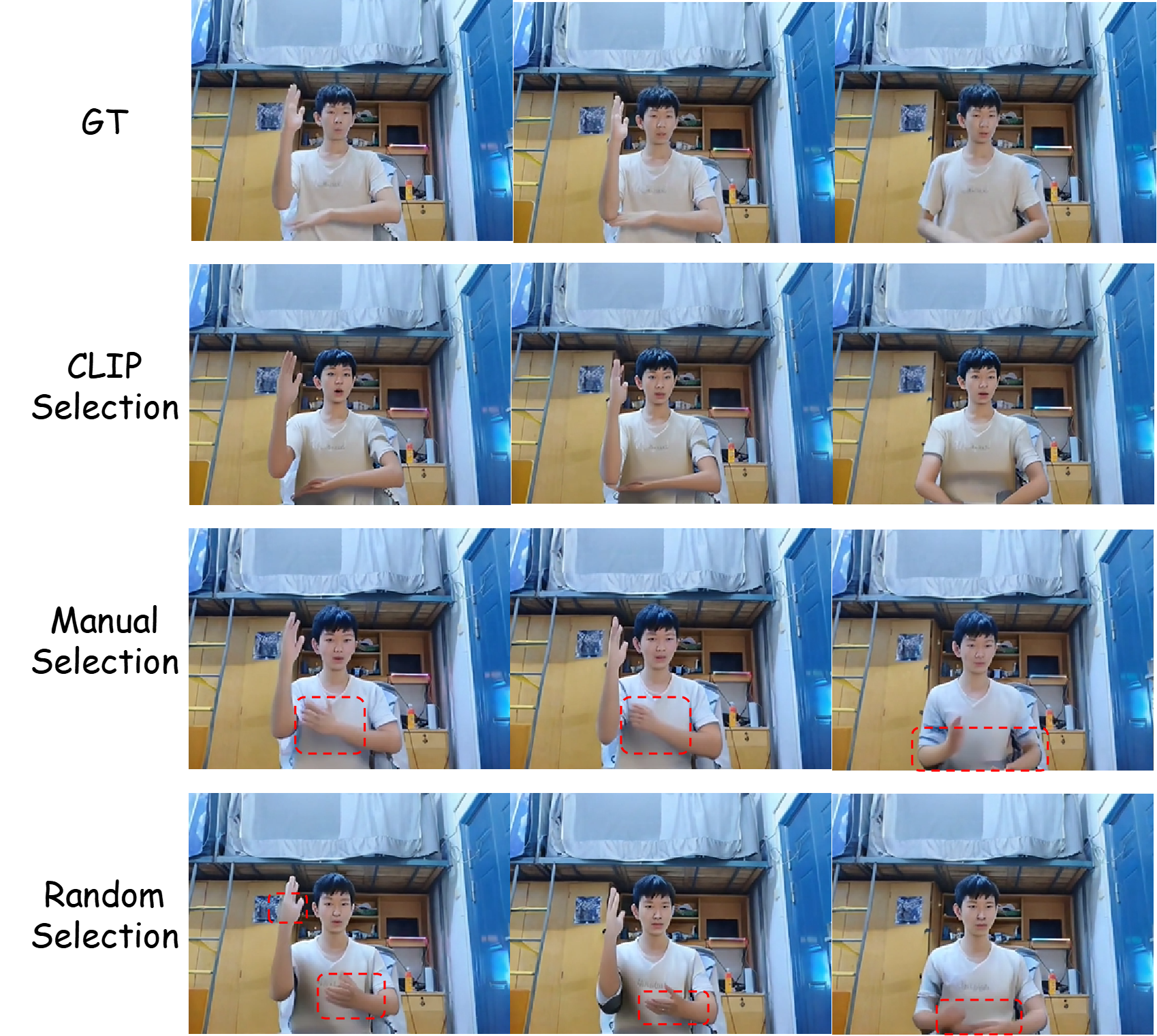}
    \caption{Qualitative results of different synthetic data selection methods. It demonstrates that the synthetic data with high semantic similarity has a relatively high movement fidelity when tested with specific data, while manual and random selection generates more blurring and artifacts.
}
    \label{fig3}
\end{figure}

\paragraph\noindent\textbf{Metrics.}
We follow standard evaluation protocols used in prior work. For image-level quality, we report Structural Similarity Index (SSIM)~\cite{wang2004image}, Learned Perceptual Image Patch Similarity (LPIPS)~\cite{zhang2018unreasonable}, and Peak Signal-to-Noise Ratio (PSNR)~\cite{hore2010image}. For temporal and perceptual coherence, we measure Fréchet Video Distance (FVD)~\cite{unterthiner2018towards}.
To further evaluate identity preservation, we calculate the cosine similarity (CSIM) of ArcFace~\cite{deng2019arcface} identity embeddings, reflecting how well the generated identity matches the source.
\begin{figure}[h]
    \centering
    \includegraphics[width=0.5\textwidth]{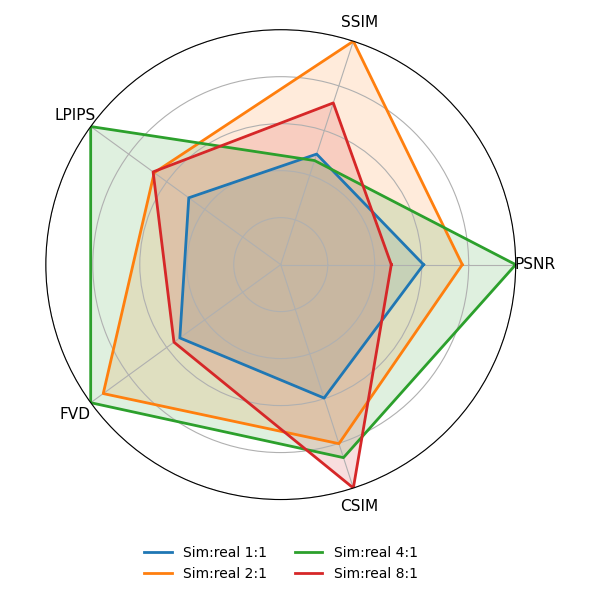}
    \caption{Radar chart comparison of different sim:real data ratios across five evaluation metrics. All metrics are normalized to [0,1] for fair comparison, sim:real=0:1 values used as normalization min, where larger values indicate better performance. Specifically, LPIPS and FVD are inverted after normalization to align their direction with other metrics (i.e., lower original values indicate better quality).
}
    \label{fig4}
\end{figure}

\vspace{-0.1in}\paragraph\noindent\textbf{Effectiveness of Synthetic Fine-Tuning.} Table~\ref{tab1} reports the quantitative results of synthetic w/o fine-Tuning on a pretrained baseline. The results show that fine-tuning the pretrained model on the synthetic dataset consistently improves generation quality across multiple metrics. This demonstrates that synthetic data can provide complementary information that real data alone may not capture—such as rare poses, complex motion transitions, or underrepresented lighting conditions. By introducing such controllable and diverse variations, fine-tuning with synthetic data effectively enhances the model’s robustness and fidelity, indicating that synthetic data serves as a valuable means to fully exploit data diversity and further boost overall model performance.
\begin{table*}[]
\centering
\caption{Comparison between models w/o finetune on synthetic data.}
\begin{tabular}{llllll}
\hline
   & PSNR↑ & SSIM↑ & LPIPS↓ & FVD↓ & ID-Sim↑ \\ \hline
Baseline & 20.0446 & 0.7219 & 0.1781 & 8.7054 & 0.4322 \\
Finetuned & \textbf{20.7764} & \textbf{0.7220} & \textbf{0.1727} & \textbf{7.1540} & \textbf{0.4666} \\ \hline
\end{tabular}
\label{tab1}
\end{table*}

\begin{table*}[]
\centering
\caption{Quantitative results of models training on data with scaling synthetic-to-real ratios.}
\begin{tabular}{cllllll}
\hline
Sim:real distribution & PSNR↑ & SSIM↑ & LPIPS↓ & FVD↓ & CSIM↑ \\ \hline
0 : 1    & 18.1346          & 0.6935           & 0.2257          & 12.9291         & 0.5042          \\
1 : 1  & 18.8544          & 0.7123          & 0.2092          & 10.6103         & 0.5911          \\
2 : 1  & 19.0496          & \textbf{0.7315} & 0.203           & 8.8505          & 0.6208          \\
4 : 1  & \textbf{19.3188} & 0.7112          & \textbf{0.1916}          & \textbf{8.5619} & 0.6299          \\
8 : 1 & 18.6918          & 0.721          & 0.2028 & 10.4754         & \textbf{0.6497} \\ \hline
\end{tabular}
\label{tab2}
\end{table*}

\begin{table}[h]
\centering
\caption{Results of synthetic data added using different methods for the specified test video}
\begin{tabular}{lllll}
\hline
Selection & PSNR↑ & SSIM↑ & LPIPS↓ & CSIM↑ \\ \hline
Random & 23.489 & 0.8165 & 0.1278 & 0.302 \\
Manual & 24.3274 & 0.8183 & 0.1266 & 0.3023 \\
CLIP-sim & \textbf{24.5244} & \textbf{0.8245} & \textbf{0.1224} & \textbf{0.3446} \\ \hline
\end{tabular}
\vspace{4pt} % 调整间距
\label{tab3}
\end{table}

\vspace{-0.1in}\paragraph\noindent\textbf{The Trade-off of Scaling Synthetic Data.} Table~\ref{tab2} presents the quantitative results of models training on data with scaling synthetic-to-real ratios, evaluated on the SpeakerVid-5M dataset, comparing models trained with different ratios of synthetic to real data. The radar chart of the experimental results is shown in Figure~\ref{fig4}. The results reveal a general upward trend in model performance as the proportion of synthetic data increases exponentially. This observation indicates that synthetic data can, to a certain extent, play a comparable role to real data in improving model generalization and visual quality.

However, the improvement is not strictly monotonic. The inclusion of certain synthetic samples contributes positively to model performance, whereas excessive or low-quality synthetic data fails to yield further gains. This instability can be attributed to the presence of a persistent Sim2Real gap, where discrepancies between synthetic and real data distributions limit the transferability of learned representations. Motivated by this observation, we conducted subsequent to explore how selective filtering of synthetic data could mitigate this gap and identify subsets of synthetic samples that are truly beneficial for model training.

\paragraph\noindent\textbf{Discussions on Targeted Synthetic Data Selection.} Table~\ref{tab3} summarizes the quantitative results of the performance of models trained on synthetic data selected by various methods. We compare three strategies for selecting synthetic data: random selection, semantic selection based on CLIP similarity, and manual human selection. The qualitative results are shown in Figure~\ref{fig3}. The results reveal that CLIP-based semantic selection outperforms both random and manual selection strategies in terms of motion smoothness, temporal coherence, and identity preservation.

This can be explained by the fact that CLIP embeddings capture high-level semantic consistency between samples.
Filtering synthetic data by CLIP similarity effectively aligns the synthetic data distribution with the target real data distribution, reducing domain discrepancy and ensuring that the model receives semantically relevant supervision signals during training. This finding suggests that automated semantic filtering offers an efficient and scalable alternative to manual selection, particularly when large-scale human annotation is impractical. More importantly, this experiment indicates that when encountering rare identities or unseen motion dynamics, specifically generating semantically related synthetic samples serves as an efficient approach to improve task-specific performance and alleviate the Sim2Real discrepancy. It further highlights the potential of targeted synthetic data selection as an effective means of narrowing the Sim2Real gap and enhancing downstream model performance in real-world video generation scenarios.

\section{Conclusion}
\label{sec:conclusion}

This work provides the first systematic exploration of synthetic data in controllable human video generation. We propose a novel and powerful unified diffusion-based framework for human controllable video generation, which serves as a platform for a series of comprehensive and pioneering experiments on synthetic data augmentation. We show that synthetic data can effectively enhance the performance of controllable video generation models, and when properly integrated, it complements real data by improving motion realism, temporal consistency, and identity preservation. Our analysis further demonstrates that targeted selection of synthetic samples can effectively mitigate the Sim2Real gap, offering a scalable alternative to manual curation. Overall, our findings highlight the practical value of synthetic data for building data-efficient and generalizable generative models and provide valuable insights for future research on leveraging synthetic data in large-scale video generation tasks.
{
    \small
    \bibliographystyle{ieeenat_fullname}
    \bibliography{main}
}


\end{document}